\documentclass[letterpaper]{article} 
\usepackage{aaai2026}  
\usepackage{times}  
\usepackage{helvet}  
\usepackage{courier}  
\usepackage[hyphens]{url}  
\usepackage{graphicx} 
\usepackage{natbib}  
\usepackage{caption} 
\usepackage{bbding}
\usepackage{pifont}
\usepackage{multirow}
\usepackage{graphicx}
\usepackage{booktabs}
\usepackage{colortbl}
\usepackage{tikz}
\usepackage{xspace}

\newcommand{\system}{\textsc{Human2Robot}\xspace}

\usepackage{tcolorbox}
\definecolor{lightblue}{HTML}{18282e}
\definecolor{lighterblue}{HTML}{f2fafd}  
\definecolor{gred}{rgb}{0.859,0.267,0.216}
\definecolor{ggreen}{rgb}{0.059,0.616,0.345}
\newtcolorbox{abox}{colback=lighterblue,colframe=lightblue, width=\linewidth}
\usepackage{amssymb}
\usepackage{amsmath} 
\usepackage{caption}
\usepackage{subcaption}
\usepackage{algorithm}
\usepackage{algorithmic}
\usepackage{footmisc}
\usepackage{newfloat}
\usepackage{listings}
\DeclareCaptionStyle{ruled}{labelfont=normalfont,labelsep=colon,strut=off} 
\floatstyle{ruled}
\newfloat{listing}{tb}{lst}{}
\floatname{listing}{Listing}
\title{Human2Robot: Learning Robot Actions from Paired Human-Robot Videos}
\author{
Sicheng Xie$^{1,2}$\thanks{Equal Contribution.}~\hspace{5pt}
Haidong Cao$^{1}$\footnotemark[1]~\hspace{5pt}
Zejia Weng$^{1}$~\hspace{5pt}
Zhen Xing$^{1}$~\hspace{5pt}
Haoran Chen$^{1}$~\hspace{5pt}\\
Shiwei Shen$^{1}$~\hspace{5pt}
Jiaqi Leng$^{1}$~\hspace{5pt}
Zuxuan Wu$^{1,2}$\thanks{Corresponding Author.}\hspace{5pt}
Yu-Gang Jiang$^{1}$~\hspace{5pt}
}
\vspace{0.1in}
\affiliations{
{$^1$Shanghai Key Lab of Intell. Info. Processing, School of CS, Fudan University} \\
{$^2$Shanghai Innovation Institute}
}

\usepackage{bibentry}

\begin{document}
\maketitle

\begin{abstract}
Distilling knowledge from human demonstrations is a promising way for robots to learn and act. Existing methods, which often rely on coarsely-aligned video pairs, are typically constrained to learning global or task-level features. As a result, they tend to neglect the fine-grained frame-level dynamics required for complex manipulation and generalization to novel tasks. We posit that this limitation stems from a vicious circle of inadequate datasets and the methods they inspire. To break this cycle, we propose a paradigm shift that treats fine-grained human-robot alignment as a conditional video generation problem. To this end, we first introduce H\&R, a novel third-person dataset containing 2,600 episodes of precisely synchronized human and robot motions, collected using a VR teleoperation system. We then present \system, a framework designed to leverage this data. \system employs a Video Prediction Model to learn a rich and implicit representation of robot dynamics by generating robot videos from human input, which in turn guides a decoupled action decoder. Our real-world experiments demonstrate that this approach not only achieves high performance on seen tasks but also exhibits one-shot generalization to novel positions, objects, instances, and even new task categories.

\end{abstract}

\section{Introduction}
\label{sec:intro}
\begin{figure}[t]
\centering
\includegraphics[width=0.85\linewidth]{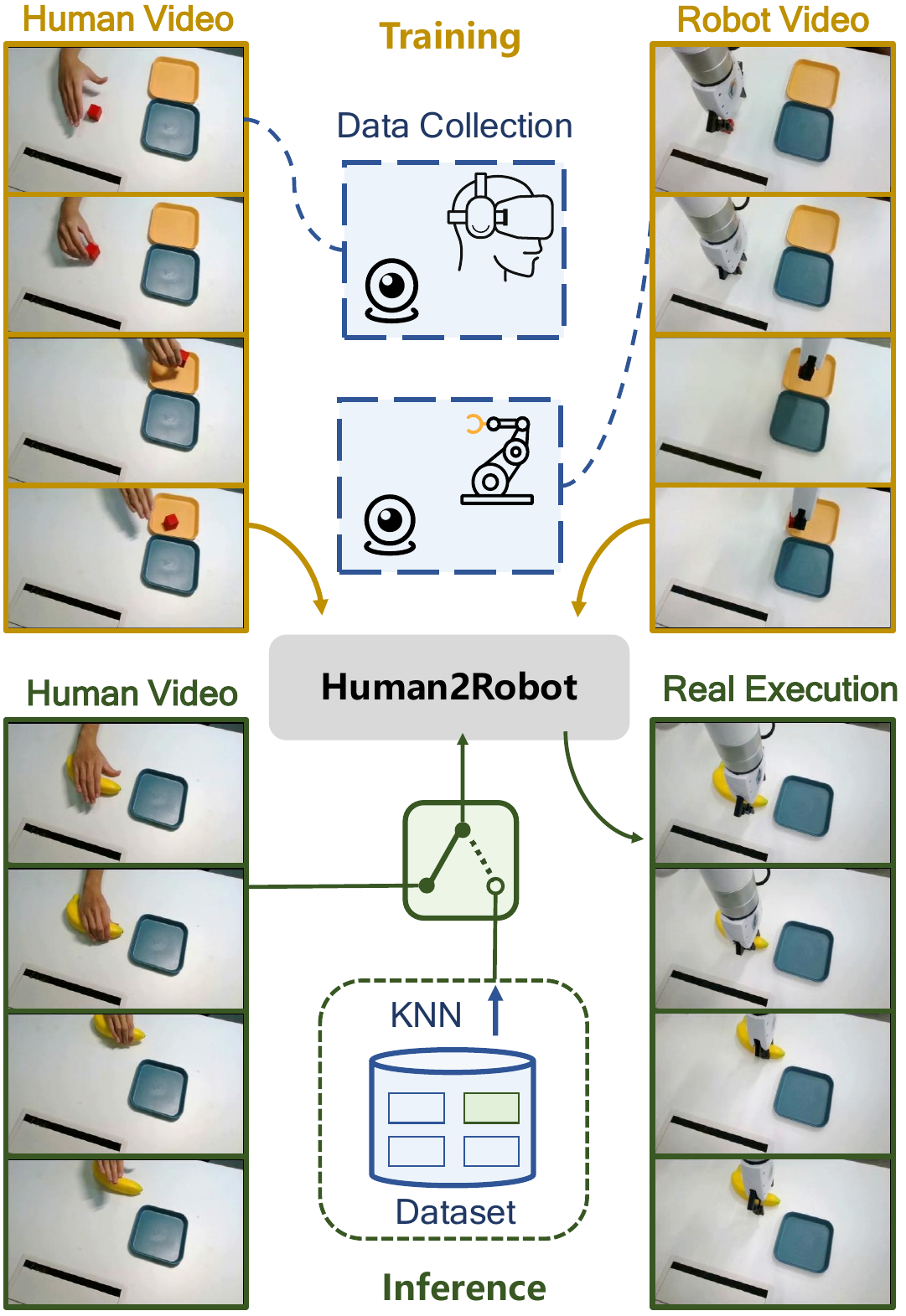}
\caption{\system: 
An human-video-conditioned policy, capable of completing seen tasks and one-shot performing unseen tasks with a single human video.
} 
\label{fig:overview}
\end{figure}

The ability to observe others, whether humans or animals, and acquire skills to solve new tasks is a fundamental reason why humans excel at tackling a wide range of complex challenges. To enable robots to assist with diverse real-world problems, it is essential that they develop a similar capability, particularly the ability to learn directly from human demonstrations. This has motivated extensive research~\cite{bahl2023affordances,srirama2024hrp,nair2022rm,wang2023mimicplay,bahl2022human,smith2019avid} on how to learn from human demonstrations effectively.

Despite recent progress, current approaches face a fundamental generalization gap. 
While they perform well on tasks encountered during training, they often fail entirely when presented with human demonstrations of unseen tasks. 
The prevailing paradigm relies on applying self-supervised~\cite{xu2023xskill} or contrastive learning~\cite{jain2024vid2robot} methods to coarsely aligned human-robot video pairs. However, the lack of fine-grained supervision in existing datasets fundamentally constrains what models can learn. Rather than uncovering detailed action-level correspondences, models are often limited to capturing global features or high-level task summaries. Consequently, many approaches extract holistic video representations using models such as the Perceiver Resampler~\cite{alayrac2022flamingo}, which compresses entire clips into fixed-length embeddings.  This severely limits their ability to model the nuanced frame-level temporal dynamics that are essential for generalization.

Furthermore, even when densely aligned data are available, models that rely on global feature matching remain fundamentally limited in their ability to capture fine-grained spatio-temporal structure. They often discard the frame-by-frame dynamics that are essential for imitating complex tasks. This limitation has created a vicious circle: the lack of fine-grained datasets gives rise to methods that are incapable of leveraging detailed supervision, while the dominance of such methods in turn discourages the development of fine-grained datasets needed to overcome this barrier. As such, we argue that to achieve true generalization where a robot can perform tasks demonstrated by a human but unseen in its own training data, we must break this cycle. This necessitates a paradigm shift in both learning methodology and data curation.

In this paper, we propose that the key to unlocking fine-grained human-robot alignment lies in video generation~\cite{vdm,gen1,align,tuneavideo,text2videozero,simda,tu2024motioneditor}. Instead of merely learning whether two videos are broadly similar or mapping task goals from the human domain to the robot domain, we aim to generate a corresponding robot video directly from a human demonstration. We posit that by training a model to predict the precise frame-by-frame evolution of a robot's movements, it learns a far richer and more temporally coherent alignment. This generative process forces the model to internalize the intricate dynamics of manipulation, enabling it to understand how a task is done, not just what is done.

To train such a model would require a dataset with densely aligned and fine-grained human-robot video pairs. Manually annotating such data is prohibitively labor-intensive. To overcome this, we leverage virtual reality teleoperation to create a novel dataset, which we call H\&R. By enhancing existing teleoperation systems with improved coordinate system matching, we achieve a seamless mapping between the operator's hand movements and the robot arm's motion. This allows us to efficiently collect a large-scale third-person dataset of 2,600 episodes, featuring perfectly synchronized videos of human hands and robot arms across 4 types of basic tasks and 6 long-horizon tasks.

Building on the proposed dataset, we introduce \system, a framework designed to leverage the proposed H\&R dataset. It employs a two-stage training process. First, we train a Video Prediction Model (VPM) built upon a pretrained Stable Diffusion model, which learns to translate a human video into a robot video. This model includes a spatial UNet for feature extraction, a behavior extractors for motion and position encoding, a spatial-temporal UNet that uses a spatial-temporal UNet architecture to explicitly model motion and temporal dynamics, generating a rich latent representation that captures core dynamics of the task. Second, we train an action decoder that conditions on the predictive representations generated by the VPM to output robot actions. 
This two-stage design effectively learn human-robot alignment and leverages the implicitly learned robot-dynamics features to guide the final policy learning.

With this carefully designed framework, \system not only excels on seen tasks but also demonstrates remarkable one-shot generalization to novel object positions, appearances, and even entirely new task types and backgrounds. Furthermore, we introduce a KNN-based inference method that allows \system to perform previously seen tasks with high precision, even without a human demonstration at test time.



In summary, our main contributions include:
\begin{itemize}
    \item We present H\&R, the first dataset featuring perfectly aligned videos of human hands and robotic arms across a variety of tasks, enabling high-fidelity learning from human demonstrations.
    \item We introduce \system, an end-to-end generative framework utilizes a two-stage training process, which excels in carefully selected tasks, even with variations in positions, appearances, instances, backgrounds and different task types.
    \item We propose a KNN+\system method, which integrates KNN for task prediction, enabling to perform tasks even without human videos as input. This further enhances the scalability and flexibility of the system. 
\end{itemize}

\section{Related Work}
\label{sec:realted_work}
\noindent\textbf{Teleoperation.} 
 Recently, VR-based methodologies~\cite{iyer2024open, ding2024bunny} have attracted considerable attention due to their cost-effectiveness, efficiency and versatility.
However, these methods focus on controlling the robot, whereas our goal of using VR is to capture perfectly aligned videos of humans and robots, which are essential for robotic imitation.
 
\noindent\textbf{Learning from Human Videos.}
Researchers have recently been attempting to leverage existing human-centric video datasets to enhance robot policy learning~\cite{liu2018imitation, smith2020avid, chen2021learning, videodex, zeng2024learning,zhang2025vlabench}. Researchers propose to learn representations from human videos to assist in task execution~\cite{xiao2022masked, wang2023mimicplay, majumdar2024searchartificialvisualcortex}. However, these approaches need strong prior knowledge and struggle to transfer to robots~\cite{nair2022rm, bahl2023affordances, bahl2022human}. Meanwhile, some human-video-conditioned methods ~\cite{gen2act, jain2024vid2robot, xu2023xskill} have focused on aligning representations across human and robot videos, they are neither efficient nor capable of generalization. In contrast, our approach uses paired data and diffusion models to achieve strong generalization capabilities. 

\noindent\textbf{Diffusion Models for Video Generation.} 
Current research has achieved remarkable performance of video generation~\cite{vdm,gen1,align,tuneavideo,text2videozero,simda,xing2023svformer,xing2025aid,xing2023vidiff,tu2024motioneditor,tu2024motionfollower,tu2024stableanimator}.
Moreover, diffusion models for video stylization~\cite{ye2025stylemaster,liu2023stylecrafter} and human image animation~\cite{hu2024animate,tu2024motioneditor} have also shown its promising potential for visual and motion transfer. 
Moreover, diffusion model not only performs impressively in generative domains but also shows promising applications in visual understanding task. Some notable approaches involve using diffusion models for object detection~\cite{diffusiondet,zhang2025diffusionad}, image segmentation~\cite{xu2023open}, and visual representation learning~\cite{yu2024representation, weng2024genrec}. 
Thus, we aim to utilize the strong capabilities of diffusion model to learn human-robot alignment and leverages the implicitly learned robot-dynamics features to guide the final policy learning.

\section{H\&R Dataset}
\label{sec:pipeline}

\subsection{Coordinate Alignment for Teleoperation}
\label{subsec:teleoperation}
Although teleoperation has become a mainstream method for data collection, no one has yet attempted to use it for collecting paired data of humans and robots. We aim to provide a feasible solution for this.

We identify two core issues. The first issue is that the coordinate system of the robotic arm is not aligned with that of the human hand, which results in a mismatch between the movements of the human hand and the robotic arm in the video. For example, the hand may move across the entire screen, while the robotic arm only moves halfway. 
This visual difference could lead to difficulties in policy learning.
The second issue is the embodiment gap between the human hand and the robotic gripper, which makes teleoperation unsuitable for certain tasks, such as screwing. 

To address the issue of coordinate system alignment, we record corresponding three points in the real world of the robotic arm and human hand. We leverage these points to establish a shared coordinate system with consistent scale for both the robotic arm and the human hand. This enables the movement range of the robotic arm to match that of the human hand, as illustrated in Figure~\ref{fig:overview}. Details on how to establish a shared coordinate system can be found in the Supplementary Materials.

However, the embodiment gap remains a challenging issue for current teleoperation system, and we leave it as future work. Therefore, at the data level, we focus on collecting pick-and-place data with human and robotic arm alignment. In our experiments, we evaluate how well the model generalizes after being trained on relatively simple data.

\subsection{Statistics of H\&R Dataset}
\begin{figure}[t]
\centering 
\includegraphics[width=\linewidth]{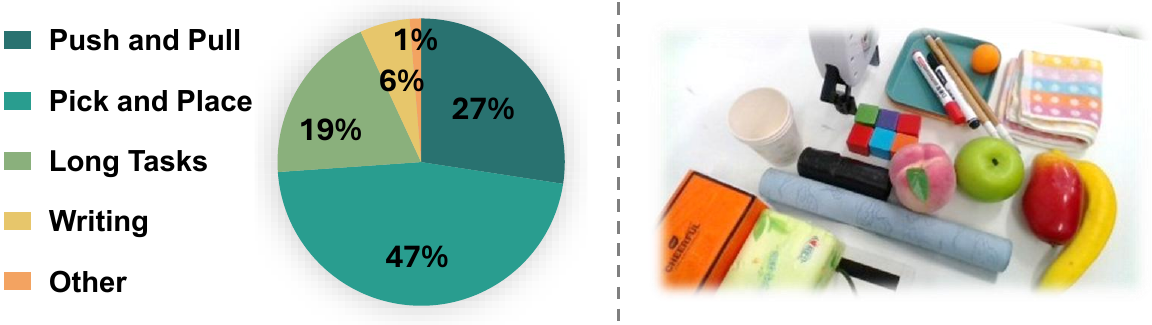}
\caption{\textbf{Dataset Overview.} \textbf{(L)} The ratio of four basic task types and long tasks. \textbf{(R)} Platform environment and the object instances used.}
\label{fig:data_overview}
\end{figure}

Based on our teleoperation method, we propose our H\&R dataset, the first dataset featuring paired human and robot videos. 
Human events vary from 4 types of basic tasks and 6 long-horizon tasks, as shown in Figure~\ref{fig:data_overview}. The whole dataset includes 2,600 episodes, and each episode contains frames ranging from 300 to 600.
To the best of our knowledge, the H\&R dataset is the first video dataset that ensures perfectly aligned video between human and robot.

\section{\system}
\label{sec:method}
\begin{figure*}[t]
\centering
\includegraphics[width=\textwidth]{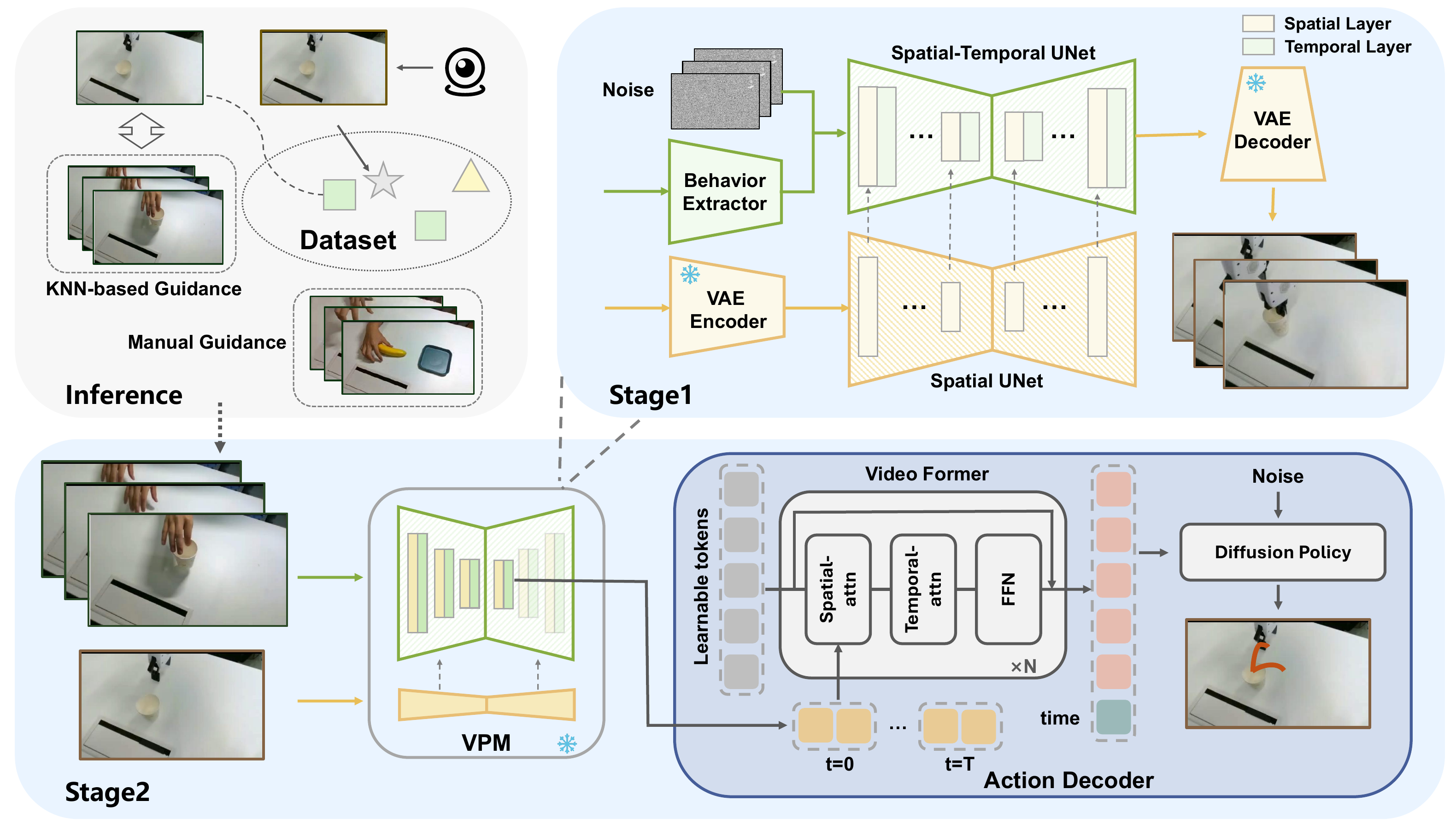}
\caption{\textbf{Architecture overview of \system.} Our approach consists of two training stages. In the first stage, we train a Video Prediction Model (VPM) to generate robotic arm videos conditioned on human videos. In the second stage, we freeze the VPM and train an action decoder to predict robot actions based on the motion features generated by the VPM.
}
\label{fig:framework}
\end{figure*}

In this section, we describe the two-stage training process of our \system. First, we explore training a VPM to generate robot videos based on human videos, implicitly learning the corresponding robotic arm actions from human movements. Next, we design an action decoder that aggregates the predictive action representations within the VPM and outputs the corresponding robot actions.


\subsection{Stage1 : Video Prediction Model (VPM)}
\label{subsec:stage1}
We first explored the possibility of generating robot videos from human videos.
Inspired by ~\cite{hu2024animate}, our VPM explores a Spatial UNet (S-UNet in short) and a Spatial-Temporal UNet (ST-UNet in short), working collaboratively to learn from humans. In particular, the S-UNet extracts features from the robot arm, which is further fed into the ST-UNet for temporal modeling. VPM also contains a behavior extractors that estimate position and motion clues from human videos. Figure~\ref{fig:framework} gives an overview of our pipeline. Below, we introduce the architecture of VPM in detail.

\noindent\textbf{Behavior Extractor.}
Recall that we aim to learn from human demonstrations, and thus it is important to extract motion and position information from human videos. This is achieved by Behavior Extractor, which contains four convolution layers (4$\times$ kernels, with a stride of 2). 
Behavior Extractor takes the human image $o_{i}^h\ (i\in[0,T])$ or the entire video of human $o_{0:T}^h$ as inputs depending on the training stage, we will talk about it later in training section.

\noindent\textbf{Spatial UNet.} 
The S-UNet, consisting of 4 upsampling blocks and 4 downsampling blocks, extracts useful clues from the robotic arm. In particular, each block, known as the Spatial Layer (S-Layer in short), leverages self-attention to learn features tailored for robot arms, as well as cross-attention with pre-extracted CLIP embedding to incorporate semantic clues. The weights of the S-UNet are initialized from those of the Stable Diffusion~\cite{rombach2022high} to ease the learning process. The features derived by S-UNet provide a condition for predicting future frames.

While S-UNet introduces a similar number of parameters as the denoising UNet, it only extracts features once during the entire process, unlike diffusion-based video generation where each frame undergoes denoising multiple times. Therefore, it does not significantly increase computational overhead during inference.

\noindent\textbf{Spatial-Temporal UNet.} 
The ST-UNet aims to predict future frames, requiring the model to learn temporal dynamics. Thus, each block of the ST-UNet has a S-Layer which is the same as the one in S-UNet, followed by an additional Temporal Layer (T-Layer in short). More specifically, the ST-UNet takes the features from the behavior extractors and noise latent as inputs. In addition, each S-Layer in ST-UNet also concatenates the features of the corresponding layer of the S-UNet, exploring the robot arm as references for prediction. The T-Layer focuses on temporal modeling so as to produce high-fidelity future frames.

\noindent\textbf{VAE Encoder and Decoder.} 
\system builds upon Stable Diffusion~\cite{rombach2022high}, which consists of an encoder and a decoder. The encoder turns images into latent embeddings for fast denoising, and the decoder maps the latents back to images. During training, both the encoder and decoder are kept frozen.

\noindent\textbf{Training strategy.}
\label{sec:training}
To generate higher-quality videos, the VPM training process consists of two stages. In the first stage, we focus on image generation to acquire basic generative capabilities. Next, we train it to generate videos.

In the first stage, by taking the first frame of robot $o^r_0$, the first frame of human $o^h_0$ and future frame of human $o^h_{i}\ (i\in[0,T])$ as inputs to predict the future frame of robot $o^r_{i}$, where $T$ refer to the maximum length of video generation. In this stage, we train the S-UNet, the Behavior Extractors, and the ST-UNet without the temporal layers. The S-Layer in both the ST-UNet and S-UNet are instantiated using the powerful open-source Stable Diffusion model (SD) ~\cite{rombach2022high}, while the Behavior Extractors are initialized with Gaussian weights, except for the final projection layer, which utilizes zero initialized convolution.
In the second stage, we concentrate the training effort on the temporal layer to generate video. In this stage, we take a 30-frame segment of human video $o^h_{0:T}$ and the first frame of robot $o^r_0$ as inputs to predict the future robot observation $o^r_{0:T}$. The entire human video segment is first passed into Behavior Extractor, which is subsequently added to noise latent and fed to our ST-UNet. The weights of the VAE Encoder and Decoder, as well as the CLIP image encoder are frozen all the time.

The optimization objectives for this stage are as follows:
\begin{equation}
    L_G = \mathop{\mathbb{E}_{z_{ts},c,\epsilon,ts}}(||\epsilon - \epsilon_\theta(z_{ts},c,ts)||_2^2),
\end{equation}
the parameters of the formula represent the latent state $z_{ts}$ at time step $ts$, which is obtained from ST-UNet. The conditioning variable $c$ includes the first frame $o^r_0$ of the robot and a human video $o^h_{0:T}$. $\epsilon$ denotes the added noise, and $\epsilon_\theta(z_{ts},c,ts)$ represents the model's predicted noise.

\subsection{Stage2 : Action Decoding.}
After the first stage of training, our VPM model is able to visually predict the corresponding robotic arm actions based on human movements. However, fully denoising an entire video is time-consuming, with most of the time spent on reconstructing image-level details, which are unrelated to manipulation. Increasingly, recent research~\cite{hu2024video,wen2024vidman,zhu2025unified} has shown that the features of a generative model after a single denoising step already contain sufficient motion features to guide the action head's planning. Inspired by these works, we treat the pretrained VPM as a vision encoder to extract the predicted robotic arm action information. This means adding intensity-invariant noise (t = K) to the human video and using the first denoised result as the feature. As shown in the Figure~\ref{fig:step1}, we can see that the one-step denoising already contains a lot of action and position information.

Additionally, previous work~\cite{weng2024genrec,xiang2023denoising}  has emphasized that the upsampling layers of diffusion models often contain more effective information. Therefore, we use the outputs of the VPM’s upsampling layers as features; more precisely, we use those from the first upsampling layer.

In summary, during this stage, we freeze the parameters of the VPM and treat it as a visual encoder to train the subsequent action decoder. We use the first upsampling layer output $F_\text{VPM}$, after the initial denoising step, as the action prior features for the action decoder.

\noindent\textbf{Action Decoder.}
Inspired by recent video conditioned works~\cite{hu2024video,luo2025learning}, our action decoder consists of two parts: the Video Former~\cite{blattmann2023align} and the Diffusion Policy~\cite{chi2023diffusion}. 

The Video Former uses a learnable token $Q$ to aggregate the video features $F_\text{VPM}$ into a feature $F_\text{VF} \in \mathbb{R}^{n\times L}$ with a fixed-length n and specified channel dimension L. Formally, this branch can be expressed as follows:
\begin{equation}
    F_\text{VF} = \text{FFN}(\text{Temp-Attn}(\text{Spat-Attn}(Q,F_{\text{VPM}}))).
\end{equation}

We incorporate the aggregated representation $F_\text{VF}$ into the diffusion transformer blocks through cross-attention layers. The diffusion policy seeks to reconstruct the original actions $a_0$ from the noisy actions $a_k = \sqrt{\overline\beta_{k}} a_0 + \sqrt{1 - \overline\beta_{k}} \epsilon$, where $\epsilon$ represents white noise,  and $\overline{\beta_{k}}$ is the noisy coefficient at step $k$. The optimization objectives for diffusion policy are as follows:
\begin{equation}
    L_A = \mathop{\mathbb{E}_{a_{k},F_{\text{VF}},\epsilon,k}}(||\epsilon - \epsilon_\phi(a_{k},F_\text{VF},k)||_2^2).
\end{equation}

\subsection{KNN + \system.}
\label{subsec:knn}
To avoid the need to explicitly provide human demonstration videos for seen task, we use a k-nearest neighbors (KNN) approach to identify the most probable task for the current scene. We retrieve the human demonstration video corresponding to the closest matching features as the conditioning input to guide the task execution. Specifically, we use DINOv2~\cite{oquab2023dinov2} and CLIP~\cite{radford2021learning} as feature extractors to capture features from the first frame of each robotic arm video in the training set. During prediction, we select the $n$ closest features based on the current environment, and the episode with the most frequent and closest match is chosen as the conditioning input, which is depicted in the Inference section of Figure~\ref{fig:framework}.

\section{Experiments}
\label{sec:experiments}
\begin{figure*}[t]
\centering
\includegraphics[width=\textwidth]{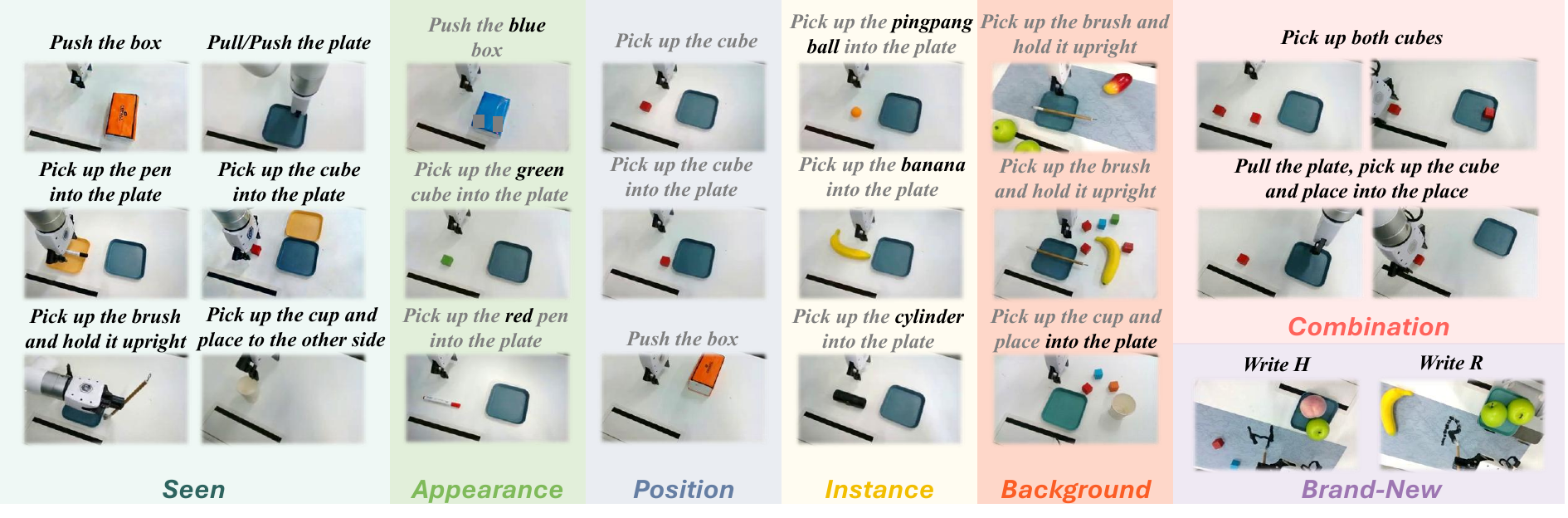}
\caption{{Task overview.
We train the models on seen tasks and test them on different generalization ability level.
}
}
\label{fig:task_overview}
\end{figure*}
\begin{table*}[t]
\centering
\begin{minipage}{0.60\linewidth}  
    \centering
    \resizebox{\linewidth}{!}{
    \begin{tabular}{l|cccc}
    \toprule
             & Push \& Pull & Pick \& Place & Rotation & Average  \\ 
    \midrule
    \textit{DP~\cite{chi2023diffusion}} & 50         & 20         & 15      &     28        \\ 
    \textit{XSkill~\cite{xu2023xskill}} & 70         & 40         & 50     &   53           \\ 
    \textit{VPP~\cite{hu2024video}} &     95     &     70     &    75   &       80        \\ 
    \midrule
    \textit{Action Decoder w. Human} &   50       &   10       &   10    &     23        \\ 
    \textit{\system w/o. Pretrain} &    20      &     10     &   0    &     10        \\ 
    \textit{\system w. KNN} & 90         & 75         & 80      &        82 \\ 
    \textit{\textbf{\system(ours)}}  & \textbf{100} & \textbf{90} & \textbf{95}  & \textbf{95}\\ 
    \bottomrule
    \end{tabular}}
    \caption{Multi-task success rate on basic tasks. Each task is evaluated with 20 trails.}
    \label{tab:main}
\end{minipage}%
\hfill  
\begin{minipage}{0.38\linewidth}
    \centering
    \resizebox{\linewidth}{!}{
    \begin{tabular}{l|cccc}
    \toprule
             Generalization  &  XSkill &  VPP & H2R(ours)  \\ 
    \midrule
    \textit{Appearance} & 0         &    80      & \textbf{100}            \\ 
    \textit{Position} & 20         &    50      & \textbf{80}     \\ 
    \textit{Instance}  & 0 & 0 & \textbf{70} \\ 
    \textit{Background}  & 0 & 0 & \textbf{80} \\ 
    \textit{Combination}  & 0 & 0 & \textbf{50} \\ 
    \textit{Brand-New}  & 0 & 0 & \textbf{70} \\ 
    \bottomrule
    \end{tabular}}
    \caption{Generalization success rate. Each task is evaluated with 20 trails.}
    \label{tab:generalization}
\end{minipage}
\end{table*}

\subsection{Experimental Setups}
\noindent\textbf{Task Definition.}
As mentioned in section~\ref{subsec:teleoperation}, even with our adjustments, current teleoperation systems still cannot collect data for difficult tasks, such as screwing, when paired human and robot arm videos are required. Therefore, we mainly focus training and testing on pick-and-place tasks. While individual tasks are relatively simple, we aim to demonstrate the generalization ability of our model after training on such tasks. This goal aligns with the spirit of learning from humans, that is, to learn from simple tasks and then generalize to more complex or even new tasks.

In addition to testing on tasks from the training set, we also test on many unseen tasks, including variations in appearance, position, instance, background, task combinations, and entirely new tasks. The task scenarios and their corresponding descriptions are shown in the Figure~\ref{fig:task_overview}.



\noindent\textbf{Baselines.} 
Since \system is characterized by learning from humans and video generation pretraining, we compare it with the following baselines:
\begin{itemize}
    \item \textbf{Diffusion Policy~\cite{chi2023diffusion}:} A action learning policy using action diffusers with CLIP language conditions. For simplicity, we refer to it as DP.
    \item \textbf{XSkill~\cite{xu2023xskill}:} A human-video-conditioned policy through self-supervised learning.
    \item \textbf{Video Prediction Policy~\cite{hu2024video}:} A language-conditioned policy that uses video prediction for pretraining. For simplicity, we refer to it as VPP.
    \item \textbf{Action Decoder w. Human:} Condition the action decoder of \system on human videos instead of the features extracted by VPM. The human videos are processed by ResNet18~\cite{he2016deep} and used as input to action decoder.
    \item \textbf{\system w/o. Pretrain:} \system without the video generation pretraining in Section~\ref{subsec:stage1}.
    \item \textbf{\system w. KNN:} \system with our KNN method proposed in Section~\ref{subsec:knn}, which enables to perform tasks without explicit demonstrations.
\end{itemize}

\noindent\textbf{\system Training Details.} 
As mentioned in Section~\ref{sec:method}, we used a two-stage training method. In the first stage, we trained a video prediction model, focusing on generating robotic arm videos based on human videos. During this stage, we pre-trained on 2,600 task videos, including some long tasks, such as picking up two blocks. Training for the first stage took 3 days using 4 NVIDIA A100 GPUs. In the second stage, we trained \system on simple tasks, as the seen tasks shown in the Figure~\ref{fig:task_overview}. This took about 6 hours using 8 NVIDIA A100 GPUs. However, for the writing task mentioned in Section~\ref{subsec:generalization}, we train \system solely on play data of writing for about 6 hours using 8 NVIDIA A100 GPUs. It is worth noting that the play data of writing here consists of random movements, without actually writing any characters.

\subsection{Main results}
\noindent\textbf{Quantitative Results.} 
The comparison on the basic tasks are show in Table~\ref{tab:main}. 
However, the DP baseline appears to converge primarily on push and pull tasks and performs poorly on the other tasks. Although XSkill conditions on human videos, it merely treats them as task labels. As a result, it can complete seen tasks but does not fully exploit the information in the human videos, leading to unstable performance and success rates that are 30–50\% lower than \system. VPP also employs video‑generation pretraining and attains success rates close to those of \system. Nevertheless, because VPP is language‑conditioned—whereas \system is video‑conditioned, providing richer, fine‑grained motion cues—\system still leads by 10–20 percentage points in success rate. Overall, our proposed \system achieves the highest success rate across all tasks, highlighting the benefits of conditioning on human videos coupled with video‑generation pretraining.

\noindent\textbf{KNN Results.} As shown in Table~\ref{tab:main}, \system with KNN outperforms all other baselines across all tasks, demonstrating that even without direct demonstrations, \system can still achieve strong performance. However, compared to \system, the KNN method shows a 10–20\% decrease in success rate, which we consider to be within an acceptable range. Overall, these competitive results demonstrate that \system is both efficient and accurate on seen tasks.

\noindent\textbf{Visualizations of Predictive Representations.}
\begin{figure*}[t]
\centering
\includegraphics[width=\linewidth]{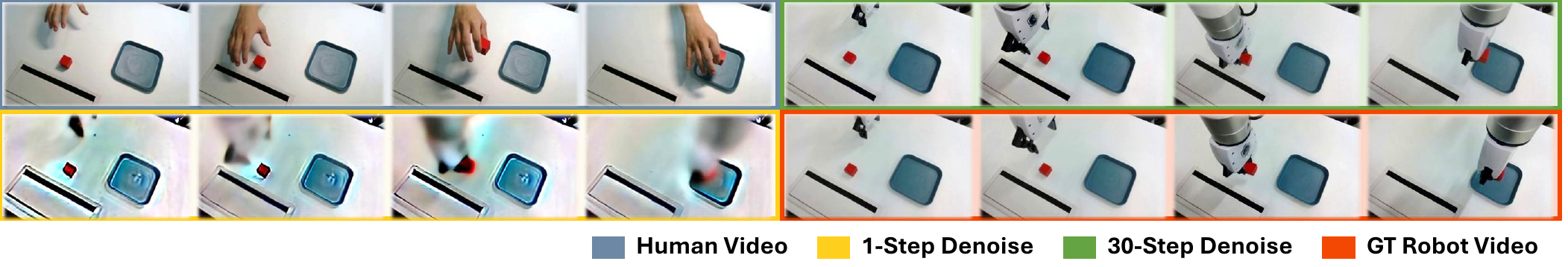}
\caption{Visualization of VPM results. We can observe that that a 1-step denoised result already contains sufficient motion information for downstream tasks. In addition, the 30-step (fully denoised) result is very close to the GT Robot video, demonstrating the effective design of our Video Prediction Model (VPM).}
\label{fig:step1}
\vspace{-3ex}
\end{figure*}

Since we repurpose the video prediction model as a visual encoder and extract predictive representations with a single forward pass, we examine the quality of these representations. In Figure~\ref{fig:step1}, we visualize the ground‑truth future alongside 1‑step predictions and 30‑step denoised outputs. The visualizations show that a 1-step denoised result already contains sufficient motion information for downstream tasks, validating the soundness of our approach. 
In addition, the 30-step (fully denoised) result  is very close to the GT Robot video, demonstrating the effective design of our VPM.


\subsection{Ablation Study}
\label{subsec:ablation}

\noindent\textbf{Effectiveness of VPM.}
Since our H\&R dataset contains paired videos of humans and robotic arms, it naturally includes paired human observations and the corresponding robot actions. Therefore, we design $\textit{Action Decoder w. Human}$ to predict robot actions directly from human videos. Specifically, we use the human video $o^h_{0:T}$ as a conditioning input to the action decoder to predict the corresponding robotic actions $a_{0:T}$.

However, in our experiments, this approach produced highly jittery executions, and the model was insensitive to grasping motions in the human videos, frequently failing to complete grasps. As a result, the average success rate was only 23\%. We attribute this to the complexity of human motions: without any prior for learning the correspondence, it is difficult to infer the correct mapping from human videos. By contrast, \system leverages video generation to learn the action correspondence, providing the downstream action decoder with a more reliable motion prior.

\noindent\textbf{Effectiveness of Video Pretraining.}
We designed \system $\textit{w/o. Pretraining}$ to assess the contribution of video‑generation pretraining. In this variant, we initialize the VPM with the initial parameters described in Section~\ref{subsec:stage1}, freeze the VPM, and train only the downstream action decoder. As shown in the table, this approach is almost incapable of completing tasks: it achieves 20\% success on the simplest push‑and‑pull tasks and only 10\% on pick‑and‑place. 

We attribute this poor performance to Stable Diffusion initialization fails to extract task‑relevant features and, in effect, injects additional noise into the observations—thereby degrading the policy. By contrast, the pretrained \system delivers strong performance, demonstrating the effectiveness of video pretraining.

\subsection{Generalization}
\label{subsec:generalization}
To evaluate generalization ability, we test not only basic generalization types—position, appearance, instance, and background—but also stronger changes: combined and brand-new tasks. As shown in Table~\ref{tab:generalization}, compared with the baselines, \system not only achieves substantially higher metrics on position and appearance, but also maintains solid performance in the other four settings where XSkill and VPP fail. We attribute this generalization to two factors: (1) unlike the self‑training approach of XSkill, our method leverages paired human–robot data from H\&R, enabling direct learning of the correspondence between human hands and robot arms; and (2) unlike VPP, our policy is human-video‑conditioned rather than language‑conditioned, so human videos supply richer, fine‑grained motion cues, allowing strong generalization even when training on simple tasks and environments.

\noindent\textbf{Appearance Generalization.} 
We tested the generalization ability on objects with different colors, materials, and textures. \system and VPP maintained their core capabilities without being affected, achieving success rates of 100\% and 80\%, respectively. However, XSkill could barely complete the tasks, showing limited generalization.

\noindent\textbf{Position Generalization.} 
Due to small variations in object placement within the dataset, all three models exhibit some degree of positional generalization. However, when the positional shift is large, XSkill achieves only a 20\% success rate, VPP reaches only 50\%, whereas \system succeeds on 80\% of the tested positions.

\noindent\textbf{Instance Generalization.} 
We tested the generalization ability on different instances. During training, all the models only encountered data for picking up blocks and pens. We tested whether the models could generalize to picking up objects such as ping pong balls and bananas. Due to the different shapes of the instances, their placement positions also varied slightly, adding complexity to the task. \system still achieved a 70\% success rate on this task, while the other baselines were unable to complete it.

\noindent\textbf{Background Generalization.} 
We also added many irrelevant objects and introduced unseen backgrounds to test the generalization ability. We found that both baselines were unable to make correct predictions under these conditions, while \system was still able to predict correctly, achieving a success rate of 80\%.

\noindent\textbf{Task Combination.} 
We believe that a model with strong generalization ability should not only be able to complete tasks it has seen before, but also be capable of handling unseen tasks. At the beginning, we think such a model should have the ability to learn from each short task and then be able to complete long tasks composed of these short tasks. The challenge in this task lies in the fact that tasks from the previous stage may affect the execution of the subsequent stage. For example, we designed a task that involves pulling the plate and placing the cube onto it. Since the plate is moved, the placement position of the block changes, introducing significant difficulty. Therefore, only \system, which has access to the corresponding human video, is capable of completing these tasks.

\noindent\textbf{Brand-New Task.} 
Teaching a robot to write is a challenging problem for current learning-based approaches. The core issue lies in the vast number of characters, with writing each character being a brand-new task. We cannot teach the model to write each character individually, so it must possess strong generalization capabilities. To demonstrate the performance of \system in writing, we created a special dataset where each trajectory involves random movements on a desktop without writing any specific characters. We used this data to train the models to learn the correspondence between human and robots, and then during inference, we instructed the models to write brand-new characters, such as "H" or "R." Experimental results show that only \system can learn the correspondence from meaningless data and complete the task of writing character.

\section{Conclusion}
\label{sec:conclusion}
In this paper, we use VR teleoperation systems to collect perfectly paired human-robot data, and create the H\&R dataset. We then present \system, which leverages a video prediction model (VPM) for human-robot alignment, and the robot-dynamics features captured turn out to be effective in guiding action decoding. Our evaluations demonstrate that \system excels in seen tasks and unseen tasks.
\subsubsection{Acknowledgments.}
This work was supported by the Science and Technology Commission of Shanghai Municipality(No. 24511103100).

\appendix
\section{System Setup Details}
\label{sec:detail_system_setup}
\begin{figure*}[t]
    \centering
    \begin{subfigure}{0.4\textwidth}
        \centering
        \includegraphics[width=\textwidth]{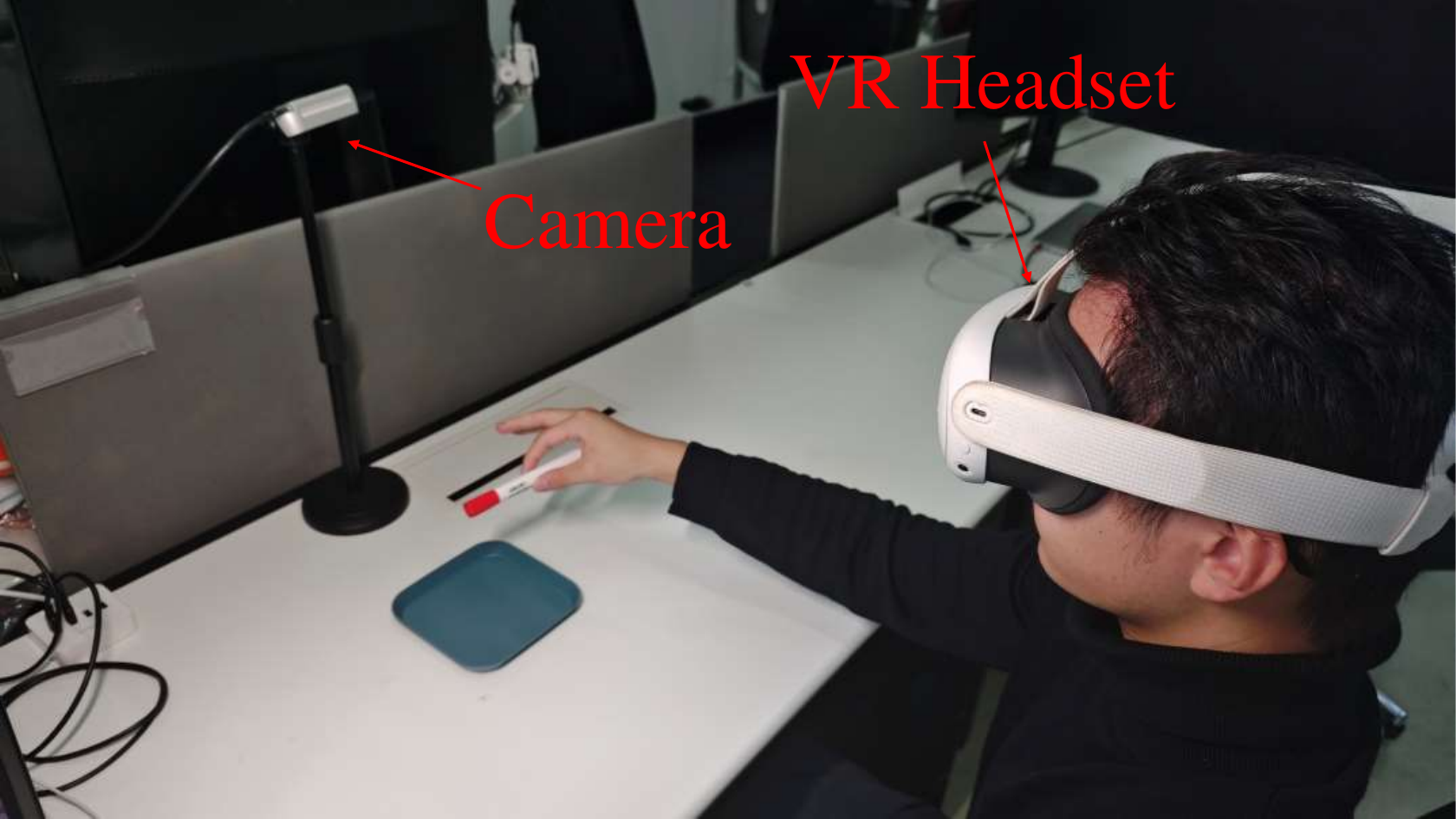}
        \caption{Human Operating Space.}
        \label{fig:human_operating_space}
    \end{subfigure}
    \hspace{0.1cm}
    \begin{subfigure}{0.4\textwidth}
        \centering
        \includegraphics[width=\textwidth]{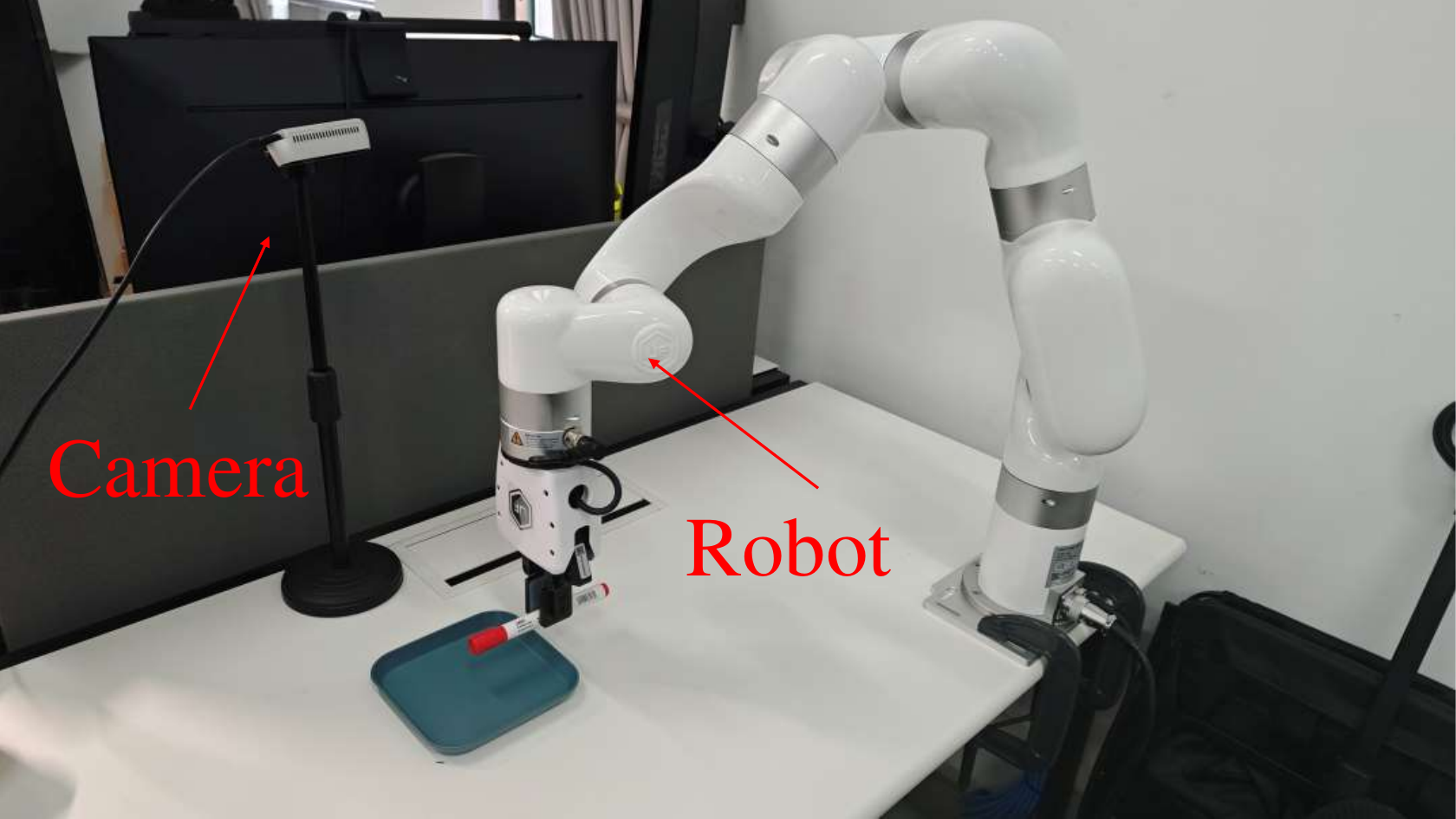}
        \caption{Robot Operating Space.}
        \label{fig:robot_operating_space}
    \end{subfigure}
    \begin{subfigure}{0.9\textwidth}
        \centering
        \includegraphics[width=\textwidth]{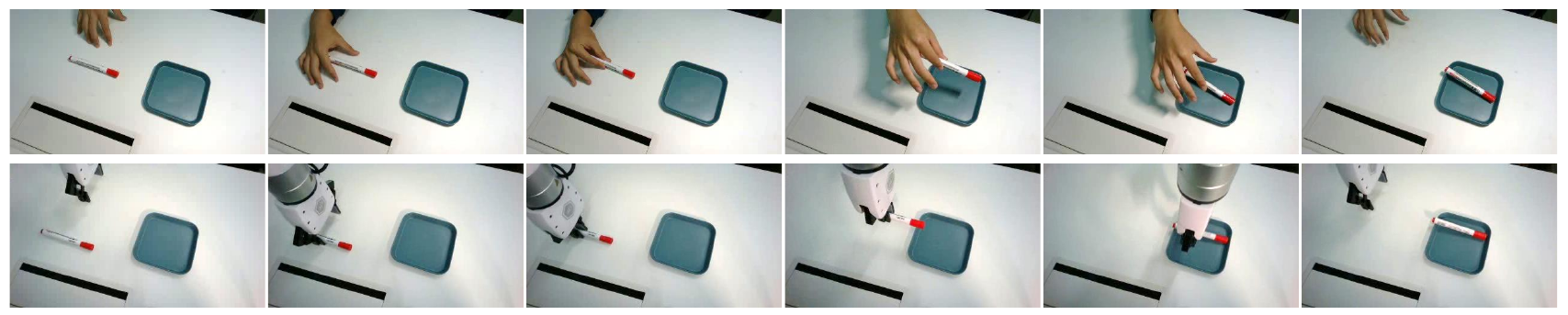}
        \caption{Data demonstration.}
        \label{fig:data_detailed_demo}
    \end{subfigure}
\caption{The operating spaces of human and robot}
\label{fig:operating_space}
\end{figure*}

The H\&R system is equipped with state-of-the-art technology specifically chosen to facilitate seamless and precise interactions between human operators and robotic mechanisms, including a Meta Quest 3 VR headset, a 7-DoF xArm robotic arm, and two Intel Realsense D435 cameras as shown in Figure~\ref{fig:human_operating_space}~\ref{fig:robot_operating_space}.

The Meta Quest 3 VR headset is selected for its advanced gesture recognition capability and affordable cost. This gesture recognition capability is integral to our data collection pipeline, providing an efficient and user-friendly way to enhance the natural interaction between human operators and the robotic system.
The Xarm robotic arm, which features seven degrees of freedom, provides the flexibility and precision required for complex and varied tasks. This arm is integral to replicating human hand movements accurately, ensuring that the physical actions translated through the VR system are executed precisely by the robotic counterpart.
Visual data collection is handled by two Intel Realsense D435 cameras strategically positioned to monitor both human and robot actions. These cameras operate at a resolution of $240\times424$ pixels, capturing images at 30Hz. The dual-camera setup captures the synchronized movements of both the human and robotic arms. With the integration of the hardware components, our system aims to achieve a critical objective: 

\begin{abox}
Producing a paired dataset where each data point reflects a seamless synchronization of human hand action and robotic arm response, captured synchronously from identical viewpoints.
\end{abox}

Toward this goal, we build our H\&R system to develop software and establish data collection protocols. We build the control system with OpenTeach~\cite{iyer2024open}, connecting different hardware components to enable the operator to perform manual tasks and simultaneously monitor the robotic arm’s actions through the VR headset display. Specifically, we set up two similar environments for human operation and robotic arm operation respectively, as shown in Figure~\ref{fig:operating_space}, and one camera is placed in each operating space. 
When the user wears the VR headset and performs hand movements, the headset transmits the coordinates of these movements to the backend, where they are retargeted to control the robotic arm, enabling synchronized actions.
Section~\ref{sec:detail_data_collection} illustrates more details about the retargeting process. Here, one of the biggest challenges lies in the inconsistency of the scale and the orientation between the two coordinate systems. In the following, we mainly illustrate our implementation for achieving the coordinate alignment.



\section{Data Collection Details}
\label{sec:detail_data_collection}

While collecting data, the operating spaces are shown in Figure~\ref{fig:operating_space}. 
During the preparation phase, we will record the positional information of four points within the robot's operating space: three anchor points and the initial position of the robotic arm. The starting position of the robotic arm is recorded and detailed anchor configurations are provided in Section~\ref{sec:alignmet}. Operators are required to wear headsets to perform operations throughout the data collection process. The robotic arm will not follow the hand's movements until the operator sets three anchor points in their workspace by putting hand on each anchor for a brief period. Our program utilizes six anchor points—three for the human and three for the robot—to align the coordinate systems of both sides, as detailed in Section~\ref{sec:alignmet}. Since the coordinate system initiated by the headset may vary slightly each time, it is necessary to make a few adjustments to the positions of objects in the environment before starting the actual data collection. 
During the data collection process, we will use the right hand for demonstration, while the left hand will control the start and end of recording by pressing the keyboard.
\subsection{Coordinate System Alignment}
\label{sec:alignmet}
In this section, we provide a detailed explanation of how we align the position and rotation of the coordinate systems for the VR headset and the robotic arm.
\begin{figure}[h]
\centering
\includegraphics[width=0.7\linewidth]{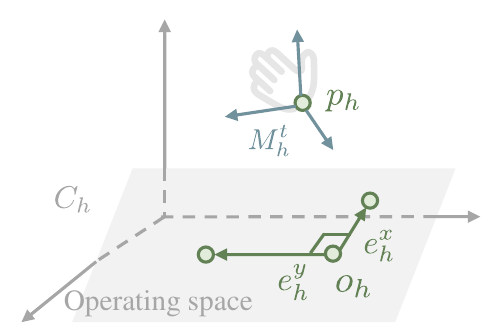}
\caption{VR headset Coordinate System.}
\label{fig:pipeline:alignment}
\end{figure}

\noindent\textbf{Position alignment.}
\label{para:position}
We utilized the introduced anchor points to align the coordinate systems. As illustrated in Figure~\ref{fig:pipeline:alignment}, we select three anchors within the operating space and derive two mutually perpendicular vectors, $e_h^x$ and $e_h^y$ for the human side, and $e_r^x$ and $e_r^y$ for the robot side. These are accompanied by the coordinates of the bottom-right anchor point, $o_h$ and $o_r$, and are subject to the following constraints in the real world, $|e_h^x|_w=|e_r^x|_w$, $|e_h^y|_w=|e_r^y|_w$ and $ e_h^x \cdot e_h^y = e_r^x \cdot e_r^y = 0$, where $|\cdot|_w$ indicates the real-world distance. From these vectors, we calculate the unit vectors $e_h^z$ and $e_r^z$, which are perpendicular to $e_h^x$ and $e_h^y$ respectively.

Then, given any hand coordinate $p_h$, the corresponding robotic arm coordinate $p_r$ can be computed as:
\begin{equation}
p_r = o_r + \mu_x e_r^x + \mu_y e_r^y + \eta \mu_z e_r^z,
\end{equation}
where $\eta$ is a hyperparameter that represents the scaling factor of motion amplitude along the z-axis. The value of $\mu$ can be computed as follows:
\begin{equation}
\mu_i = \frac{e_h^i\cdot (p_h - o_h)}{|e_h^i|^2}, i \in \{ x, y, z \}.
\end{equation}
\noindent\textbf{Rotation alignment.}
After aligning the positions, additional alignment is necessary for the orientations. Our goal is to ensure that the relative rotations of the human hand and the robotic arm, relative to their initial states, remain consistent. 

To synchronize the rotation of the robotic arm with the human hand, we establish a 3D coordinate system similar to the approach taken by OpenTeach\cite{iyer2024open}. This coordinate system enables us to compute the rotation matrix $M_h^t$ for the human hand at time $t$. Additionally, we derive the rotation matrix $M_r^t$ for the robotic arm at the same time $t$, using the robot's state parameters such as pitch, yaw, and roll. The rotation matrix $M_h^t$ and $M_r^t$ at the time $t$ can be expressed in terms of the rotation matrices $M_h^0$ and $M_r^0$ at time $t = 0$ as:
\begin{equation}
    M_h^t = M_h^0 R_h^t,
    \label{equ:rotation_h}
\end{equation}
\begin{equation}
    M_r^t = M_r^0 R_r^t,
    \label{equ:rotation_r}
\end{equation}
where $R_h^t$ and $R_h^t$ represent the rotation relative to the coordinate system $M_h^0$ and $M_r^0$ respectively. Recall that our goal is to ensure the relative rotations are the same, which allows us to derive the following equation:
\begin{equation}
    R_r^t = P^{-1}R_h^tP,
    \label{equ:rotation_relative}
\end{equation}
where $P$ is the transformation matrix defined from the basis of the coordinate system $M_r^0$ to that of $M_h^0$.

Organize Equation~\ref{equ:rotation_h}, ~\ref{equ:rotation_r}, and~\ref{equ:rotation_relative}, and the target rotation matrix of the robotic arm can be calculated as follows:
\begin{equation}
M_r^{i} = M_r^0 P^{-1} (M_h^{0})^{-1} M_h^{i} P,
\end{equation}

\section{Supplementary Experiment Results}
\begin{figure*}[h]
\centering
\includegraphics[width=\textwidth]{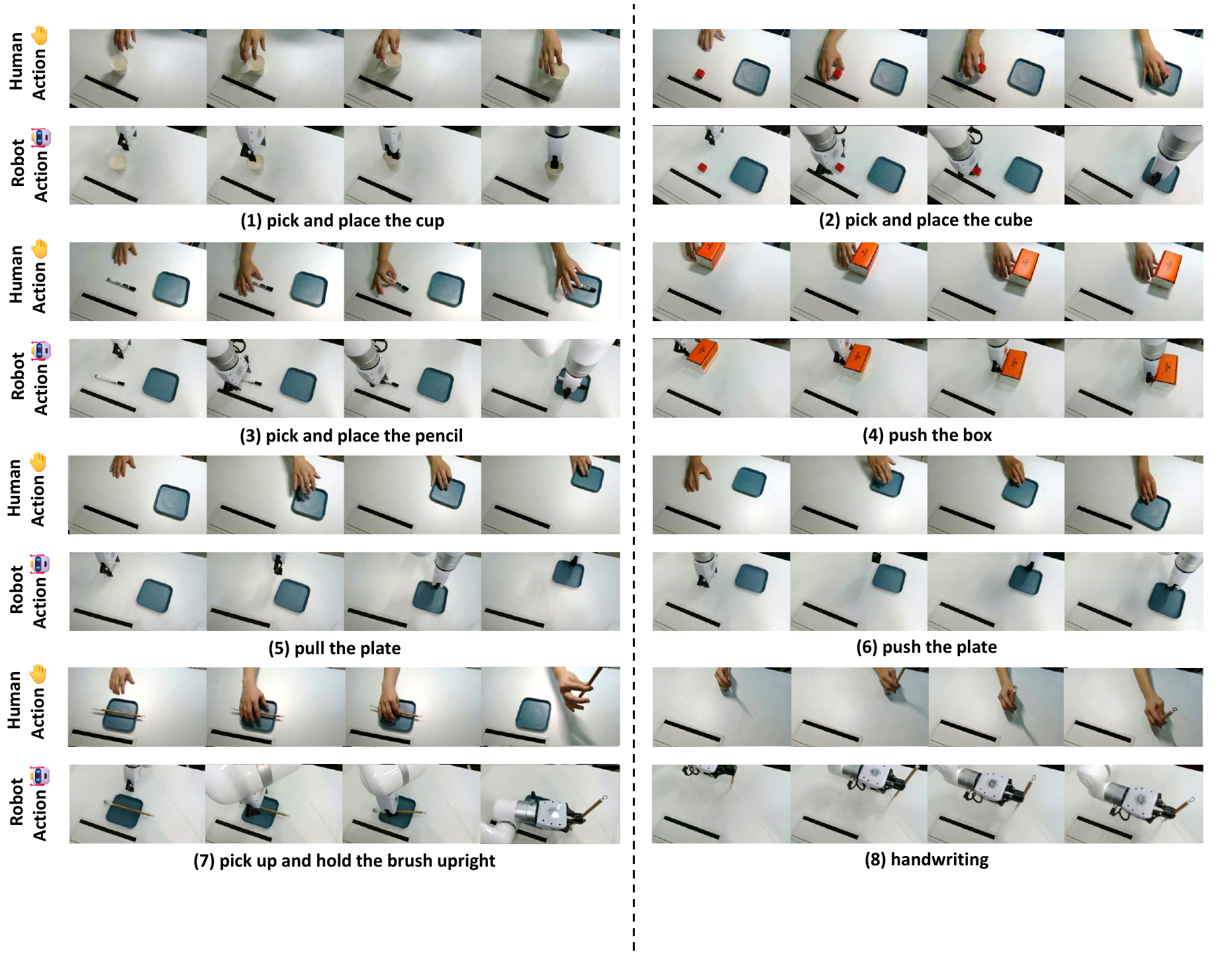}
\caption{\textbf{Overview of the dataset for base tasks.} }
\label{fig:dataset_overview}
\end{figure*}

\subsection{System stability}
\noindent\textbf{Accuracy.} 
There are two main factors affecting control accuracy: one is the precision of the data source, i.e., the accuracy of hand gesture recognition, and the other is the accuracy loss caused by mapping hand movements to the robotic arm. Our gesture recognition data comes from the Meta Quest 3, and we found that under sufficient lighting conditions, it provides highly accurate hand tracking.

A traditional mapping approach is to map the wrist position to the  end-effector position of robot. However, this creates a topological inconsistency, leading to accuracy errors. A more topologically consistent approach is to map the midpoint between the thumb and index fingertip. While this method preserves structural consistency, it introduces jitter due to the differing speeds of the thumb and index finger when grasping objects. Thus, we track the Metacarpophalangeal Joint (MCP) of the index finger, ensuring both the prevention of jitter and the maintenance of accuracy.

\noindent\textbf{Jitter Handling.}
We set the control mode to serial, making the system’s sensitivity depend on execution speed. By allowing 100–300ms latency, we trade speed for stability, reducing jitter and ensuring smoother motion.

\begin{figure*}[h]
\centering
\includegraphics[width=\textwidth]{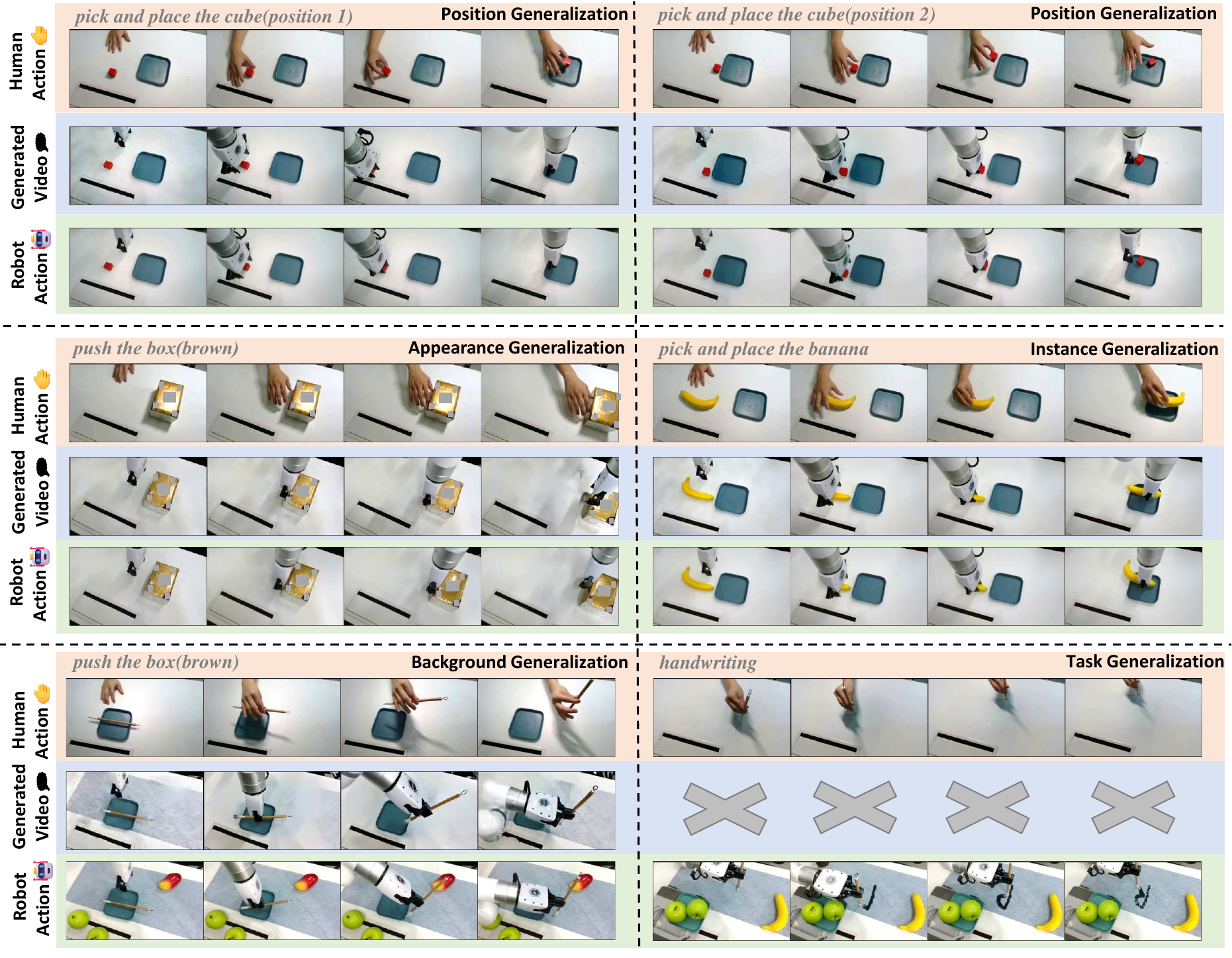}
\vspace{-1.6em}
\caption{\textbf{Visualization of generalization evaluation process.} To accelerate inference, we skip the video generation process when performing the handwriting task.}
\vspace{-1em}
\label{fig:generalization_demo}
\end{figure*}

\section{H\&R Details}
\subsection{Task Details}
Our H\&R dataset contains 8 base tasks and 6 long-horizon tasks. The definition of our tasks in our dataset is listed below.

\noindent\textbf{Base tasks.}
The visual representations of the tasks are shown in Figure~\ref{fig:dataset_overview}.
\begin{itemize}
    \item Pick and place the cup: The cup can initially be placed anywhere on the table. The goal of this task is to pick up the cup and place it in another location.
    \item Pick and place the cube: The cube comes in two colors (red and green), and it can initially be placed within a designated area. The goal of this task is to pick up the cube and place it on a plate.
    \item Pick and place pencil: The pencil comes in two colors (red and black), the goal is to pick up the pencil and place it on a plate.
    \item Push the box: The box comes in two colors (red and green), and its initial position of it is random. The goal is to push the box from left to right
    \item Pull the plate: The goal is to pull the plate from bottom to top.
    \item Push the plate: The goal is to push the plate from top to bottom.
    \item Pick up and hold the brush upright: The goal is to pick up the brush and upright it.
    \item Handwriting: Play data, write aimlessly on a blank desk.
\end{itemize}

\noindent\textbf{Long-Horizon tasks.}
These tasks are a combination of the base tasks.
\begin{itemize}
    \item Pick both cubes: There are two red cubes on the table, and the goal is to pick up each cube individually and place them onto the plate.
    \item Pick the designated cube: There is a red cube and a green cube on the table, and the goal is to pick up one of the cubes onto the plate.
    \item Pick to the designated plate: There is a red cube and two plates on the table, and the goal is to pick up the cubes onto one of the plates.
    \item Pick the cube and pull the plate: The combination of pinch a cube and pull the plate. The goal is to first pick the cube onto the plate and pull the plate.
    \item Pull the plate and pick the cube: The combination of pull the plate and pinch a cube. After pulling the plate, the cube should be placed in the plate's new position.
    \item Return to the initial position: Complete the base tasks and return to the initial position.
\end{itemize}

\subsection{Episode Details}
H\&R consists of 2,600 episodes, each containing the following information, adhering to the RT-X standards:
\begin{itemize}
    \item a paired human hand and robotic arm video containing frames between 200 and 600, with variations depending on the task.
    \item state of the robotic arm at each timestamp, including position and rotation.
    \item joint of the robotic arm velocity at each timestamp.
    \item transform matrix of the human hand, including position and rotation.
    \item position of each key point of the human hand.
    \item action of the robotic arm retargeted based on human hand data.
    \item timestamp.
\end{itemize}

\bibliography{main}

@String(CVPR= {IEEE Conf. Comput. Vis. Pattern Recog.})

@String(ICCV= {Int. Conf. Comput. Vis.})

@String(ICLR = {Int. Conf. Learn. Represent.})

@String(CVPR  = {CVPR})

@String(RSS  = {RSS})

@String(ICCV  = {ICCV})

@String(ICLR  = {ICLR})

@article{iyer2024open,
  title={Open teach: A versatile teleoperation system for robotic manipulation},
  author={Iyer, Aadhithya and Peng, Zhuoran and Dai, Yinlong and Guzey, Irmak and Haldar, Siddhant and Chintala, Soumith and Pinto, Lerrel},
  journal={arXiv preprint arXiv:2403.07870},
  year={2024}
}

@inproceedings{xu2023open,
  title={Open-vocabulary panoptic segmentation with text-to-image diffusion models},
  author={Xu, Jiarui and Liu, Sifei and Vahdat, Arash and Byeon, Wonmin and Wang, Xiaolong and De Mello, Shalini},
  booktitle={CVPR},
  year={2023}
}

@inproceedings{diffusiondet,
  title={Diffusiondet: Diffusion model for object detection},
  author={Chen, Shoufa and Sun, Peize and Song, Yibing and Luo, Ping},
  booktitle={ICCV},
  year={2023}
}

@inproceedings{simda,
  title={Simda: Simple diffusion adapter for efficient video generation},
  author={Xing, Zhen and Dai, Qi and Hu, Han and Wu, Zuxuan and Jiang, Yu-Gang},
  booktitle={CVPR},
  year={2024}
}

@inproceedings{xing2023svformer,
  title={Svformer: Semi-supervised video transformer for action recognition},
  author={Xing, Zhen and Dai, Qi and Hu, Han and Chen, Jingjing and Wu, Zuxuan and Jiang, Yu-Gang},
  booktitle={CVPR},
  year={2023}
}

@inproceedings{xing2025aid,
  title={Aid: Adapting image2video diffusion models for instruction-guided video prediction},
  author={Xing, Zhen and Dai, Qi and Weng, Zejia and Wu, Zuxuan and Jiang, Yu-Gang},
  booktitle={ICCV},
  year={2025}
}

@article{xing2023vidiff,
  title={Vidiff: Translating videos via multi-modal instructions with diffusion models},
  author={Xing, Zhen and Dai, Qi and Zhang, Zihao and Zhang, Hui and Hu, Han and Wu, Zuxuan and Jiang, Yu-Gang},
  journal={arXiv preprint arXiv:2311.18837},
  year={2023}
}

@inproceedings{hu2024animate,
  title={Animate anyone: Consistent and controllable image-to-video synthesis for character animation},
  author={Hu, Li},
  booktitle={CVPR},
  year={2024}
}

@article{ding2024bunny,
  title={Bunny-visionpro: Real-time bimanual dexterous teleoperation for imitation learning},
  author={Ding, Runyu and Qin, Yuzhe and Zhu, Jiyue and Jia, Chengzhe and Yang, Shiqi and Yang, Ruihan and Qi, Xiaojuan and Wang, Xiaolong},
  journal={arXiv preprint arXiv:2407.03162},
  year={2024}
}

@inproceedings{nair2022rm,
title={R3M: A Universal Visual Representation for Robot Manipulation},
author={Suraj Nair and Aravind Rajeswaran and Vikash Kumar and Chelsea Finn and Abhinav Gupta},
booktitle={CoRL},
year={2022},
}

@article{xiao2022masked,
  title={Masked visual pre-training for motor control},
  author={Xiao, Tete and Radosavovic, Ilija and Darrell, Trevor and Malik, Jitendra},
  journal={arXiv preprint arXiv:2203.06173},
  year={2022}
}

@inproceedings{majumdar2024searchartificialvisualcortex,
  title={Where are we in the search for an artificial visual cortex for embodied intelligence?},
  author={Majumdar, Arjun and Yadav, Karmesh and Arnaud, Sergio and Ma, Jason and Chen, Claire and Silwal, Sneha and Jain, Aryan and Berges, Vincent-Pierre and Wu, Tingfan and Vakil, Jay and others},
  booktitle={NeuIPS},
  year={2023}
}

@article{weng2024genrec,
  title={Genrec: Unifying video generation and recognition with diffusion models},
  author={Weng, Zejia and Yang, Xitong and Xing, Zhen and Wu, Zuxuan and Jiang, Yu-Gang},
  journal={arXiv preprint arXiv:2408.15241},
  year={2024}
}

@article{smith2020avid,
  title={Avid: Learning multi-stage tasks via pixel-level translation of human videos},
  author={Smith, Laura and Dhawan, Nikita and Zhang, Marvin and Abbeel, Pieter and Levine, Sergey},
  journal={arXiv preprint arXiv:1912.04443},
  year={2020}
}

@article{smith2019avid,
  title={Avid: Learning multi-stage tasks via pixel-level translation of human videos},
  author={Smith, Laura and Dhawan, Nikita and Zhang, Marvin and Abbeel, Pieter and Levine, Sergey},
  journal={arXiv preprint arXiv:1912.04443},
  year={2019}
}

@article{videodex,
    title={VideoDex: Learning Dexterity
    from Internet Videos},
    author={Shaw, Kenneth and Bahl,
    Shikhar and Pathak, Deepak},
    journal= {CoRL},
    year={2022}
  }

@article{chen2021learning,
  title={Learning generalizable robotic reward functions from" in-the-wild" human videos},
  author={Chen, Annie S and Nair, Suraj and Finn, Chelsea},
  journal={RSS},
  year={2021}
}

@article{bahl2022human,
  title={Human-to-robot imitation in the wild},
  author={Bahl, Shikhar and Gupta, Abhinav and Pathak, Deepak},
  journal={arXiv preprint arXiv:2207.09450},
  year={2022}
}

@article{oquab2023dinov2,
  title={Dinov2: Learning robust visual features without supervision},
  author={Oquab, Maxime and Darcet, Timoth{\'e}e and Moutakanni, Th{\'e}o and Vo, Huy and Szafraniec, Marc and Khalidov, Vasil and Fernandez, Pierre and Haziza, Daniel and Massa, Francisco and El-Nouby, Alaaeldin and others},
  journal={arXiv preprint arXiv:2304.07193},
  year={2023}
}

@article{wang2023mimicplay,
  title={Mimicplay: Long-horizon imitation learning by watching human play},
  author={Wang, Chen and Fan, Linxi and Sun, Jiankai and Zhang, Ruohan and Fei-Fei, Li and Xu, Danfei and Zhu, Yuke and Anandkumar, Anima},
  journal={arXiv preprint arXiv:2302.12422},
  year={2023}
}

@inproceedings{radford2021learning,
  title={Learning transferable visual models from natural language supervision},
  author={Radford, Alec and Kim, Jong Wook and Hallacy, Chris and Ramesh, Aditya and Goh, Gabriel and Agarwal, Sandhini and Sastry, Girish and Askell, Amanda and Mishkin, Pamela and Clark, Jack and others},
  booktitle={ICML},
  year={2021},
}

@article{zeng2024learning,
  title={Learning Manipulation by Predicting Interaction},
  author={Zeng, Jia and Bu, Qingwen and Wang, Bangjun and Xia, Wenke and Chen, Li and Dong, Hao and Song, Haoming and Wang, Dong and Hu, Di and Luo, Ping and others},
  journal={arXiv preprint arXiv:2406.00439},
  year={2024}
}

@article{jain2024vid2robot,
  title={Vid2robot: End-to-end video-conditioned policy learning with cross-attention transformers},
  author={Jain, Vidhi and Attarian, Maria and Joshi, Nikhil J and Wahid, Ayzaan and Driess, Danny and Vuong, Quan and Sanketi, Pannag R and Sermanet, Pierre and Welker, Stefan and Chan, Christine and others},
  journal={arXiv preprint arXiv:2403.12943},
  year={2024}
}

@inproceedings{liu2018imitation,
  title={Imitation from observation: Learning to imitate behaviors from raw video via context translation},
  author={Liu, YuXuan and Gupta, Abhishek and Abbeel, Pieter and Levine, Sergey},
  booktitle={ICRA},
  year={2018},
}

@inproceedings{bahl2023affordances,
  title={Affordances from human videos as a versatile representation for robotics},
  author={Bahl, Shikhar and Mendonca, Russell and Chen, Lili and Jain, Unnat and Pathak, Deepak},
  booktitle={CVPR},
  year={2023}
}

@article{srirama2024hrp,
  title={Hrp: Human affordances for robotic pre-training},
  author={Srirama, Mohan Kumar and Dasari, Sudeep and Bahl, Shikhar and Gupta, Abhinav},
  journal={arXiv preprint arXiv:2407.18911},
  year={2024}
}

@inproceedings{vdm,
  title={Video diffusion models},
  author={Ho, Jonathan and Salimans, Tim and Gritsenko, Alexey and Chan, William and Norouzi, Mohammad and Fleet, David J},
  booktitle={NeuIPS},
  year={2022}
}

@inproceedings{gen1,
  title={Structure and content-guided video synthesis with diffusion models},
  author={Esser, Patrick and Chiu, Johnathan and Atighehchian, Parmida and Granskog, Jonathan and Germanidis, Anastasis},
  booktitle={ICCV},
  year={2023}
}

@inproceedings{align,
  title={Align your latents: High-resolution video synthesis with latent diffusion models},
  author={Blattmann, Andreas and Rombach, Robin and Ling, Huan and Dockhorn, Tim and Kim, Seung Wook and Fidler, Sanja and Kreis, Karsten},
  booktitle={CVPR},
  year={2023}
}

@inproceedings{tuneavideo,
  title={Tune-a-video: One-shot tuning of image diffusion models for text-to-video generation},
  author={Wu, Jay Zhangjie and Ge, Yixiao and Wang, Xintao and Lei, Stan Weixian and Gu, Yuchao and Shi, Yufei and Hsu, Wynne and Shan, Ying and Qie, Xiaohu and Shou, Mike Zheng},
  booktitle={ICCV},
  year={2023}
}

@InProceedings{text2videozero,
    author    = {Khachatryan, Levon and Movsisyan, Andranik and Tadevosyan, Vahram and Henschel, Roberto and Wang, Zhangyang and Navasardyan, Shant and Shi, Humphrey},
    title     = {Text2Video-Zero: Text-to-Image Diffusion Models are Zero-Shot Video Generators},
    booktitle = {ICCV},
    year      = {2023},
}

@article{gen2act,
  title={Gen2act: Human video generation in novel scenarios enables generalizable robot manipulation},
  author={Bharadhwaj, Homanga and Dwibedi, Debidatta and Gupta, Abhinav and Tulsiani, Shubham and Doersch, Carl and Xiao, Ted and Shah, Dhruv and Xia, Fei and Sadigh, Dorsa and Kirmani, Sean},
  journal={arXiv preprint arXiv:2409.16283},
  year={2024}
}

@inproceedings{tu2024motioneditor,
  title={Motioneditor: Editing video motion via content-aware diffusion},
  author={Tu, Shuyuan and Dai, Qi and Cheng, Zhi-Qi and Hu, Han and Han, Xintong and Wu, Zuxuan and Jiang, Yu-Gang},
  booktitle={CVPR},
  year={2024}
}

@article{tu2024motionfollower,
  title={Motionfollower: Editing video motion via lightweight score-guided diffusion},
  author={Tu, Shuyuan and Dai, Qi and Zhang, Zihao and Xie, Sicheng and Cheng, Zhi-Qi and Luo, Chong and Han, Xintong and Wu, Zuxuan and Jiang, Yu-Gang},
  journal={arXiv preprint arXiv:2405.20325},
  year={2024}
}

@article{tu2024stableanimator,
  title={StableAnimator: High-Quality Identity-Preserving Human Image Animation},
  author={Tu, Shuyuan and Xing, Zhen and Han, Xintong and Cheng, Zhi-Qi and Dai, Qi and Luo, Chong and Wu, Zuxuan},
  journal={arXiv preprint arXiv:2411.17697},
  year={2024}
}

@inproceedings{rombach2022high,
  title={High-resolution image synthesis with latent diffusion models},
  author={Rombach, Robin and Blattmann, Andreas and Lorenz, Dominik and Esser, Patrick and Ommer, Bj{\"o}rn},
  booktitle={CVPR},
  year={2022}
}

@article{yu2024representation,
  title={Representation Alignment for Generation: Training Diffusion Transformers Is Easier Than You Think},
  author={Yu, Sihyun and Kwak, Sangkyung and Jang, Huiwon and Jeong, Jongheon and Huang, Jonathan and Shin, Jinwoo and Xie, Saining},
  journal={arXiv preprint arXiv:2410.06940},
  year={2024}
}

@inproceedings{he2016deep,
  title={Deep residual learning for image recognition},
  author={He, Kaiming and Zhang, Xiangyu and Ren, Shaoqing and Sun, Jian},
  booktitle={CVPR},
  year={2016}
}

@inproceedings{xu2023xskill,
  title={Xskill: Cross embodiment skill discovery},
  author={Xu, Mengda and Xu, Zhenjia and Chi, Cheng and Veloso, Manuela and Song, Shuran},
  booktitle={CoRL},
  year={2023},
}

@inproceedings{ye2025stylemaster,
  title={Stylemaster: Stylize your video with artistic generation and translation},
  author={Ye, Zixuan and Huang, Huijuan and Wang, Xintao and Wan, Pengfei and Zhang, Di and Luo, Wenhan},
  booktitle={CVPR},
  year={2025}
}

@article{liu2023stylecrafter,
  title={Stylecrafter: Enhancing stylized text-to-video generation with style adapter},
  author={Liu, Gongye and Xia, Menghan and Zhang, Yong and Chen, Haoxin and Xing, Jinbo and Wang, Yibo and Wang, Xintao and Yang, Yujiu and Shan, Ying},
  journal={arXiv preprint arXiv:2312.00330},
  year={2023}
}

@article{zhu2025unified,
  title={Unified world models: Coupling video and action diffusion for pretraining on large robotic datasets},
  author={Zhu, Chuning and Yu, Raymond and Feng, Siyuan and Burchfiel, Benjamin and Shah, Paarth and Gupta, Abhishek},
  journal={arXiv preprint arXiv:2504.02792},
  year={2025}
}

@article{hu2024video,
  title={Video prediction policy: A generalist robot policy with predictive visual representations},
  author={Hu, Yucheng and Guo, Yanjiang and Wang, Pengchao and Chen, Xiaoyu and Wang, Yen-Jen and Zhang, Jianke and Sreenath, Koushil and Lu, Chaochao and Chen, Jianyu},
  journal={arXiv preprint arXiv:2412.14803},
  year={2024}
}

@article{wen2024vidman,
  title={Vidman: Exploiting implicit dynamics from video diffusion model for effective robot manipulation},
  author={Wen, Youpeng and Lin, Junfan and Zhu, Yi and Han, Jianhua and Xu, Hang and Zhao, Shen and Liang, Xiaodan},
  journal={NeuIPS},
  year={2024}
}

@inproceedings{xiang2023denoising,
  title={Denoising diffusion autoencoders are unified self-supervised learners},
  author={Xiang, Weilai and Yang, Hongyu and Huang, Di and Wang, Yunhong},
  booktitle={ICCV},
  year={2023}
}

@inproceedings{luo2025learning,
  title={Learning Video-Conditioned Policy on Unlabelled Data with Joint Embedding Predictive Transformer},
  author={Luo, Hao and Lu, Zongqing},
  booktitle={ICLR},
  year={2025}
}

@inproceedings{blattmann2023align,
  title={Align your latents: High-resolution video synthesis with latent diffusion models},
  author={Blattmann, Andreas and Rombach, Robin and Ling, Huan and Dockhorn, Tim and Kim, Seung Wook and Fidler, Sanja and Kreis, Karsten},
  booktitle={ICCV},
  year={2023}
}

@article{chi2023diffusion,
  title={Diffusion policy: Visuomotor policy learning via action diffusion},
  author={Chi, Cheng and Xu, Zhenjia and Feng, Siyuan and Cousineau, Eric and Du, Yilun and Burchfiel, Benjamin and Tedrake, Russ and Song, Shuran},
  journal={RSS},
  year={2023},
}

@article{alayrac2022flamingo,
  title={Flamingo: a visual language model for few-shot learning},
  author={Alayrac, Jean-Baptiste and Donahue, Jeff and Luc, Pauline and Miech, Antoine and Barr, Iain and Hasson, Yana and Lenc, Karel and Mensch, Arthur and Millican, Katherine and Reynolds, Malcolm and others},
  journal={NeuIPS},
  year={2022}
}

@article{zhang2025diffusionad,
  title={DiffusionAD: Norm-guided one-step denoising diffusion for anomaly detection},
  author={Zhang, Hui and Wang, Zheng and Zeng, Dan and Wu, Zuxuan and Jiang, Yu-Gang},
  journal={TPAMI},
  year={2025},
}

@inproceedings{zhang2025vlabench,
  title={Vlabench: A large-scale benchmark for language-conditioned robotics manipulation with long-horizon reasoning tasks},
  author={Zhang, Shiduo and Xu, Zhe and Liu, Peiju and Yu, Xiaopeng and Li, Yuan and Gao, Qinghui and Fei, Zhaoye and Yin, Zhangyue and Wu, Zuxuan and Jiang, Yu-Gang and others},
  booktitle={ICCV},
  year={2025}
}

\end{document}